\documentclass[letterpaper, 10 pt, conference]{ieeeconf}  

\IEEEoverridecommandlockouts                              
\usepackage[utf8]{inputenc}
\usepackage[T1]{fontenc}

\usepackage{graphics} 
\usepackage{amsmath} 
\usepackage{amssymb}  
\usepackage{subcaption,graphicx,dblfloatfix}

\usepackage{rotating}
\usepackage{outlines}
\usepackage{multirow, multicol}
\usepackage[usenames,dvipsnames,svgnames,table,xcdraw]{xcolor}
\usepackage{booktabs}
\usepackage{hyperref}

\newcommand{\squeezeuphalf}{\vspace{-1.5mm}}
\newcommand{\squeezeup}{\vspace{-3mm}}

\allowdisplaybreaks
\title{\LARGE \bf
Uncertainty-Aware Voxel based 3D Object Detection and Tracking with von-Mises Loss
}

\author{Yuanxin Zhong$^{1}$, Minghan Zhu$^{1}$ and Huei Peng$^{1}$
\thanks{$^{1}$Yuanxin Zhong, Minghan Zhu and Huei Peng are with Mechanical Engineering, University of Michigan, Ann Arbor, USA
        {\tt\small \{zhongyx, minghanz, hpeng\} at umich.edu}}%
}

\begin{document}

\maketitle
\thispagestyle{empty}
\pagestyle{empty}

\begin{abstract}

Object detection and tracking is a key task in autonomy. Specifically, 3D object detection and tracking have been an emerging hot topic recently. Although various methods have been proposed for object detection, uncertainty in the 3D detection and tracking tasks has been less explored. Uncertainty helps us tackle the error in the perception system and improve robustness. In this paper, we propose a method for improving target tracking performance by adding uncertainty regression to the SECOND detector, which is one of the most representative algorithms of 3D object detection. Our method estimates positional and dimensional uncertainties with Gaussian Negative Log-Likelihood (NLL) Loss for estimation and introduces von-Mises NLL Loss for angular uncertainty estimation. We fed the uncertainty output into a classical object tracking framework and proved that our method increased the tracking performance compared against the vanilla tracker with constant covariance assumption.

\end{abstract}

\section{Introduction}

3D object detection and tracking system is one of the most critical module in systems including adaptive cruise control, automatic emergency braking and autonomous vehicles, where safety is critical. It's the core part of a vehicular perception system. To achieve safe driving, the detection system needs to be both precise and reliable. Precise detected position and size are close to the values of real targets, while for reliability the numbers of false positives and false negatives are desired to be lowered.

Lidar, camera and radars are common sensors used to perform object detection. Lidars provide precise range measurement, cameras provide richer and denser information of the surroundings by color and texture, and radars provided a reliable but rough estimation of object locations and velocity. Among these sensors, Lidar is the one used most in 3D object detection given it can provide a precise 3D point cloud representing the world. There are several datasets and benchmarks designed for 3D object detection, including famous ones of KITTI\cite{geiger2013vision}, Waymo\cite{sun2020scalability} and Nuscenes\cite{caesar2020nuscenes}. These datasets enable the training of deep neural network models for object detection. Many popular object detectors using the point cloud are based on point representation or volumetric representation \cite{cui2020deep}. The former methods originate from the PointNet \cite{qi2017pointnet} structure which directly consumes raw point cloud with multiple Multi-Layer Perceptrons (MLPs) to process point-wise features and some pooling operations to gather the global information. On the other hand, the volumetric representation, pioneered by VoxelNet \cite{zhou2018voxelnet} uses voxels to divide the Euclidean space and apply standard or sparse 3D convolution on the voxel grid to extract features. After the feature being extracted, object proposals will be generated from the feature map and a Non-Maximum Suppression (NMS) process is usually performed to exclude redundant object proposals.

After object proposals are generated in each time frame, object tracking is going to report the correspondence and dynamic properties of the objects, including velocity and angular velocity. A classical model-based tracking method using detected objects usually consists of three parts \cite{badue2020self}: object association, state estimator and object buffer which is in charge of initializing and discontinuing track. Object association creates a correspondence between object proposals in the current time frame and tracked targets in previous time frames, popular methods for data association include Global Nearest Neighbor (GNN) \cite{choi2013multi}, Joint Probabilistic Data Association (JPDA) \cite{fortmann1983sonar} and Multiple Hypothesis Tracking (MHT)\cite{reid1979algorithm}. State estimator updates the pose state and object parameters based on the newest object proposal, which is usually done by Bayesian Filters (e.g Kalman Filter). Many researchers have tried to apply model-based tracking algorithm on deep object detectors, such as \cite{bewley2016simple}. They usually assume constant observation noise in filtering and let it to be a parameter to tune.

In this paper, we proposed a novel von-Mises loss for regression on angular uncertainty and we created an object tracking pipeline with detection uncertainty integrated. The main contributions of this paper include
\begin{itemize}
    \item We introduced uncertainty output to 3D detection methods and improved detection performance.
    \item We proposed a novel von-Mises NLL loss for uncertainty regression on the yaw angle of object.
    \item We proposed several scoring strategies based on uncertainty output and validated their effectiveness.
    \item We showed improvement in the tracking performance by applying estimated uncertainty from detection.
\end{itemize}

The remainder of this paper is structured as follows, Section \ref{sec:review} introduces related research in the literature, Section \ref{sec:methodology} describes the methodology of uncertainty regression in our algorithm. Sections \ref{sec:experiments} show our experiments on the proposed algorithm and ablation study of our innovations. Finally comes the Section \ref{sec:conclusion} of our conclusions.

\begin{figure*}[!t]
\centering
\includegraphics[width=\textwidth]{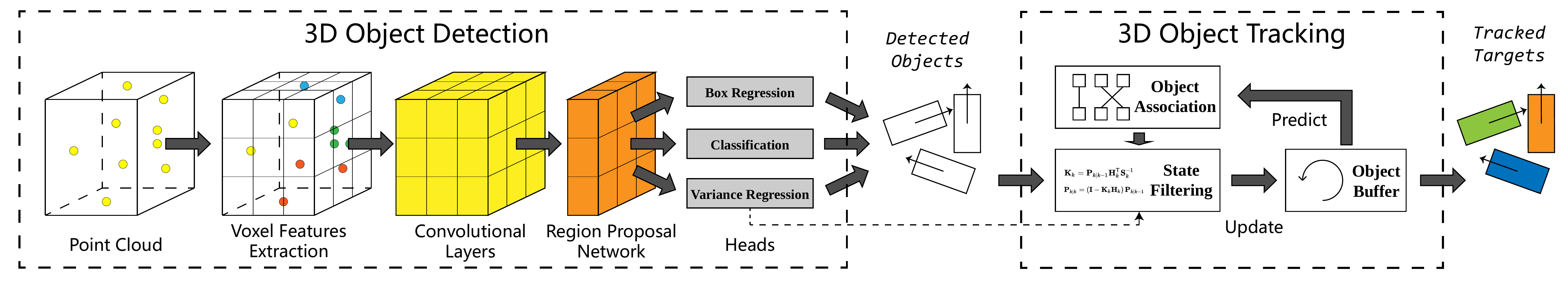}
\caption{The overall structure of proposed variance-aware detection and tracking algorithm}
\label{fig:framework}
\squeezeup
\end{figure*}

\section{Related Work}
\label{sec:review}

\subsection{3D Object Detection and Tracking}

To detect objects in 3D space, researchers have developed methods using cameras, lidars or fusion of them. 2D object detection with camera has been extensively explored \cite{ren2015faster, tan2020efficientdet, lin2017focal}, but there's less work on 3D detection with camera \cite{chen2016monocular}. The insight behind monocular and stereo 3D detection is that neural networks will learn depth information in addition to object location from the images. Image-based object detection methods can reuse most of the techniques in 2D object detection, but the lack of explicit depth information makes it hard to achieve high localization accuracy. On the other hand, since the lidar can directly provide accurate depth information, more and more works are focusing on getting 3D object proposals from the lidar point cloud. Point cloud based object detection algorithms can be put into three categories by their representation of point cloud. Methods based on PointNet \cite{qi2017pointnet} directly take raw point cloud array as input and use MLPs and some group pooling module to process point-wise information; Methods similar to or based on VoxelNet \cite{zhou2018voxelnet} use voxel grids to transform unstructured point cloud into structure grids to apply 3D convolution; Methods with projection \cite{chen2017multi} will convert the point cloud into projected images with different views, and then apply 2D convolutional networks to do detection. There are also many methods \cite{chen2017multi, sindagi2019mvx} fused the information from multi-modal sensors for 3D detection by specialized algorithm structures.

Based on whether there is a region cropping process, the object detectors are usually split into single-stage detectors and two-stage detectors. Single-stage detectors \cite{qi2017pointnet, zhou2018voxelnet, yan2018second, yang2018pixor} use processed full feature maps, while two-stage \cite{qi2018frustum, shi2019pointrcnn, chen2017multi} detectors can directly use the original point cloud with full resolution. They share a common head for box parameter regression, which makes our modification applicable to all of these methods.

\subsection{Uncertainty estimation in deep models}

Uncertainty estimation for the neural network was brought up since the Bayesian Neural Network \cite{neal2012bayesian} was proposed. BNN assigns a distribution (by adding a variance parameter) over the weights of the neural network to estimate epistemic uncertainty. Then the network can be trained with weights, bias and variance being optimized simultaneously. During the inference pass, networks with the same structure but different weights can be sampled and the variance of the result can be calculated using Monte Carlo. Besides adding distribution directly to weights, \cite{gal2016dropout} uses the Dropout layer as an alternative way to generate different networks and estimate the final posterior. 

The methods above are based on neural network sampling and the final distribution is calculated based on Monte Carlo. They suffer from the higher computation brought by evaluating the network several times to get an estimation of the uncertainty. In \cite{kendall2017uncertainties}, the authors provided a clear explanation of the difference of the aleatoric and epistemic uncertainty, and they summarized a heteroscedastic regression algorithm for direct estimation of the aleatoric uncertainty from the data, without the need of sampling network structure.

There are several previous works about applying uncertainty to object detection. \cite{le2018uncertainty} introduced two algorithms to capture aleatoric uncertainty to the SSD \cite{liu2016ssd} detector. One uses the heteroscedastic regression mentioned above and another one estimates the uncertainty from all detections proposals of the anchors. They claimed that both methods can assign large uncertainty values to false positives. In \cite{choi2019gaussian}, the author similarly added heteroscedastic regression to YOLOv3 \cite{redmon2018yolov3} and showed that the mAP of the detection network was improved by about 3\% based on their experiments. These methods all focus on 2D object detection and they didn't elaborate on how uncertainty can be practically used in downstream perception modules. We applied uncertainty regression in the 3D object detector and show the benefit of uncertainty output by using it in the object tracking problem.

\section{Uncertainty Aware Object Detection}
\label{sec:methodology}

\subsection{Detection and Tracking Framework}
The overview of the algorithm is illustrated in Figure \ref{fig:framework}.

\subsubsection{3D Detection Backbone}

SECOND\cite{yan2018second} was chosen as our detection backbone thanks to its high efficiency and regular voxel representation of point cloud. SECOND is based on the VoxelNet\cite{zhou2018voxelnet} structure, which consists of Voxel Feature Encoding layers, Convolutional Middle Layers and Region Proposal Networks. Based on that, SECOND implemented the sparse convolution on the voxels and drastically improved the computation efficiency. It also adds improvements to the parametric representation on the bounding box and data augmentation flow. Here we explain the box parameter encoding for reference of modification later in the paper:
\begin{gather}
\nonumber x_t=\frac{x_g-x_a}{d_a}, y_t=\frac{y_g-y_a}{d_a}, z_t=\frac{z_g-z_a}{h_a}, \theta_t=\theta_g-\theta_a,\\
w_t=\log\left(\frac{w_g}{w_a}\right), l_t=\log\left(\frac{l_g}{l_a}\right), h_t=\log\left(\frac{h_g}{h_a}\right)
\end{gather}

Here $x$, $y$ and $z$ are the coordinate of the bounding box center, $w$, $l$, $h$ are respectively the width, length and height of the box. $\theta$ is the yaw angle of the box (rotation around z-axis). In the formula $d=\sqrt{l^2+w^2}$ is the diagonal of the anchor box. The meanings of subscripts are $t$ for regression target, $a$ for anchor and $g$ for ground-truth values.

In the original SECOND paper, the loss on the classification branch is the Focal Loss, the losses on the box parameter regression branch are SmoothL1 Loss for position and dimension and Sine-Error Loss for the yaw angle. We follow the same box encoding and loss pattern used in SECOND, with additional branch and loss about the uncertainty of parameters. 

\begin{figure}[t]
    \centering
    \begin{subfigure}[b]{0.48\columnwidth}
        \includegraphics[width=\columnwidth]{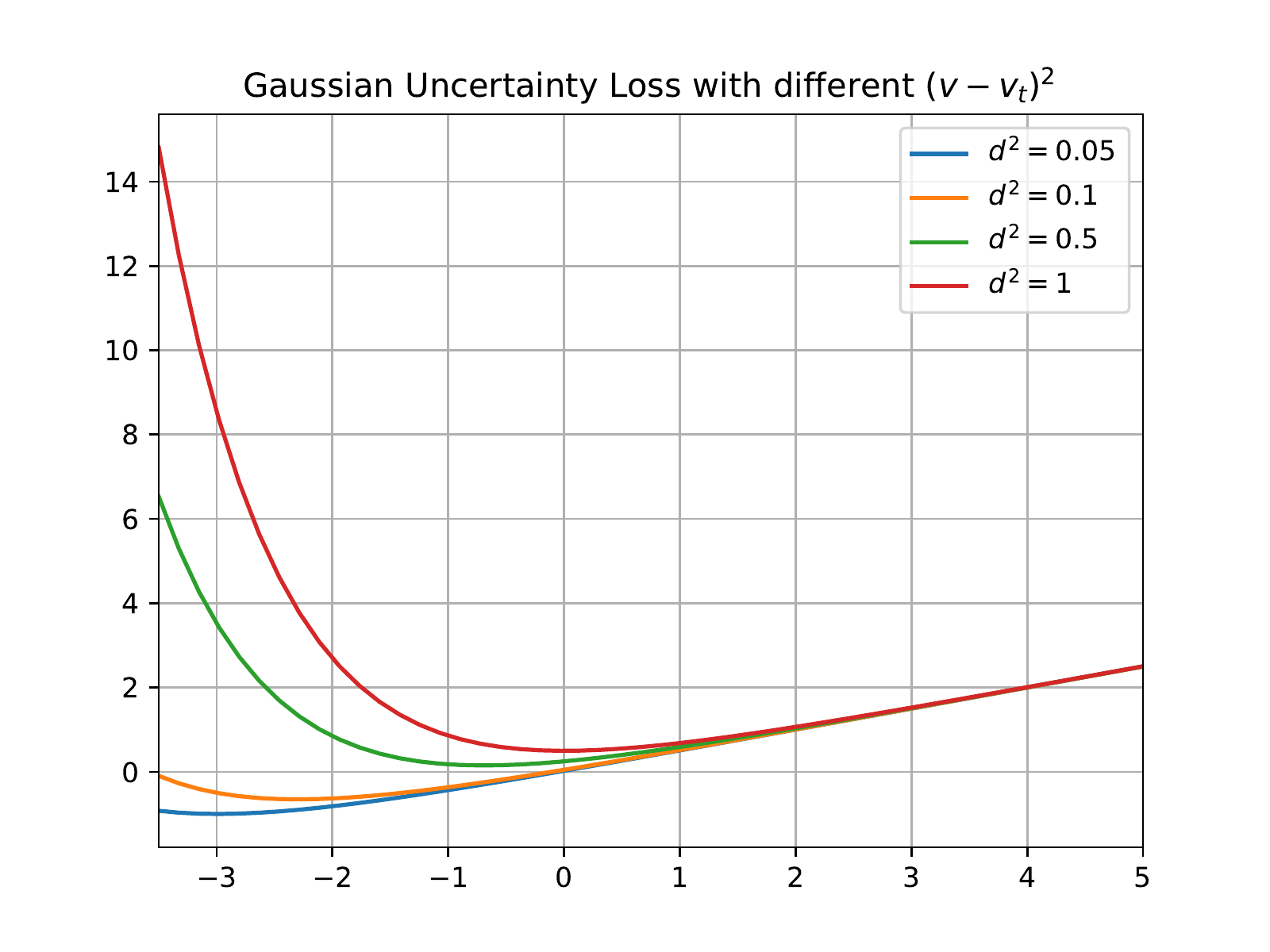}
        \caption{}
    \end{subfigure}
    \hfill
    \begin{subfigure}[b]{0.48\columnwidth}
        \includegraphics[width=\columnwidth]{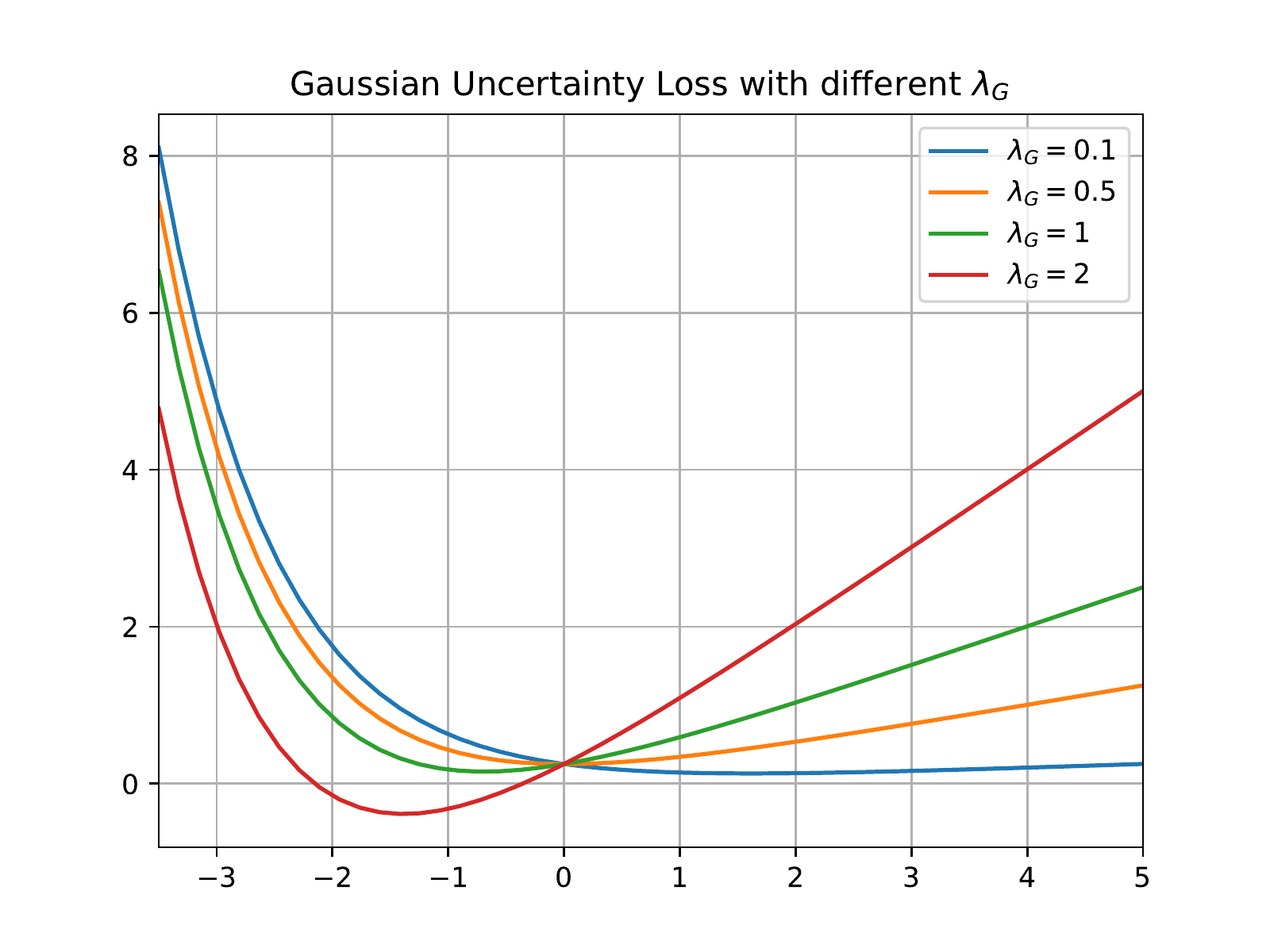}
        \caption{}
    \end{subfigure}
    \hfill
    \begin{subfigure}[b]{0.48\columnwidth}
        \includegraphics[width=\columnwidth]{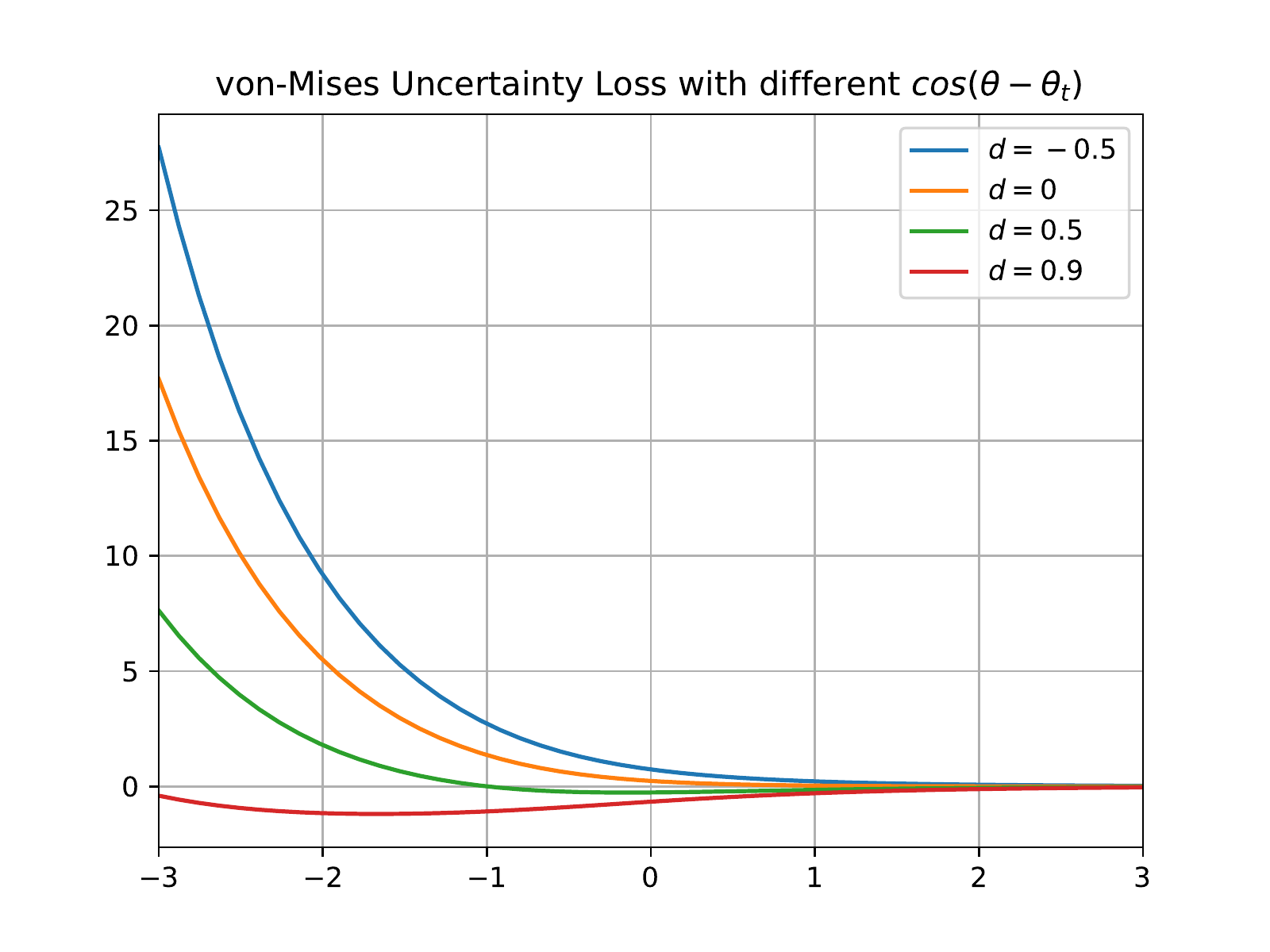}
        \caption{}
    \end{subfigure}
    \hfill
    \begin{subfigure}[b]{0.48\columnwidth}
        \includegraphics[width=\columnwidth]{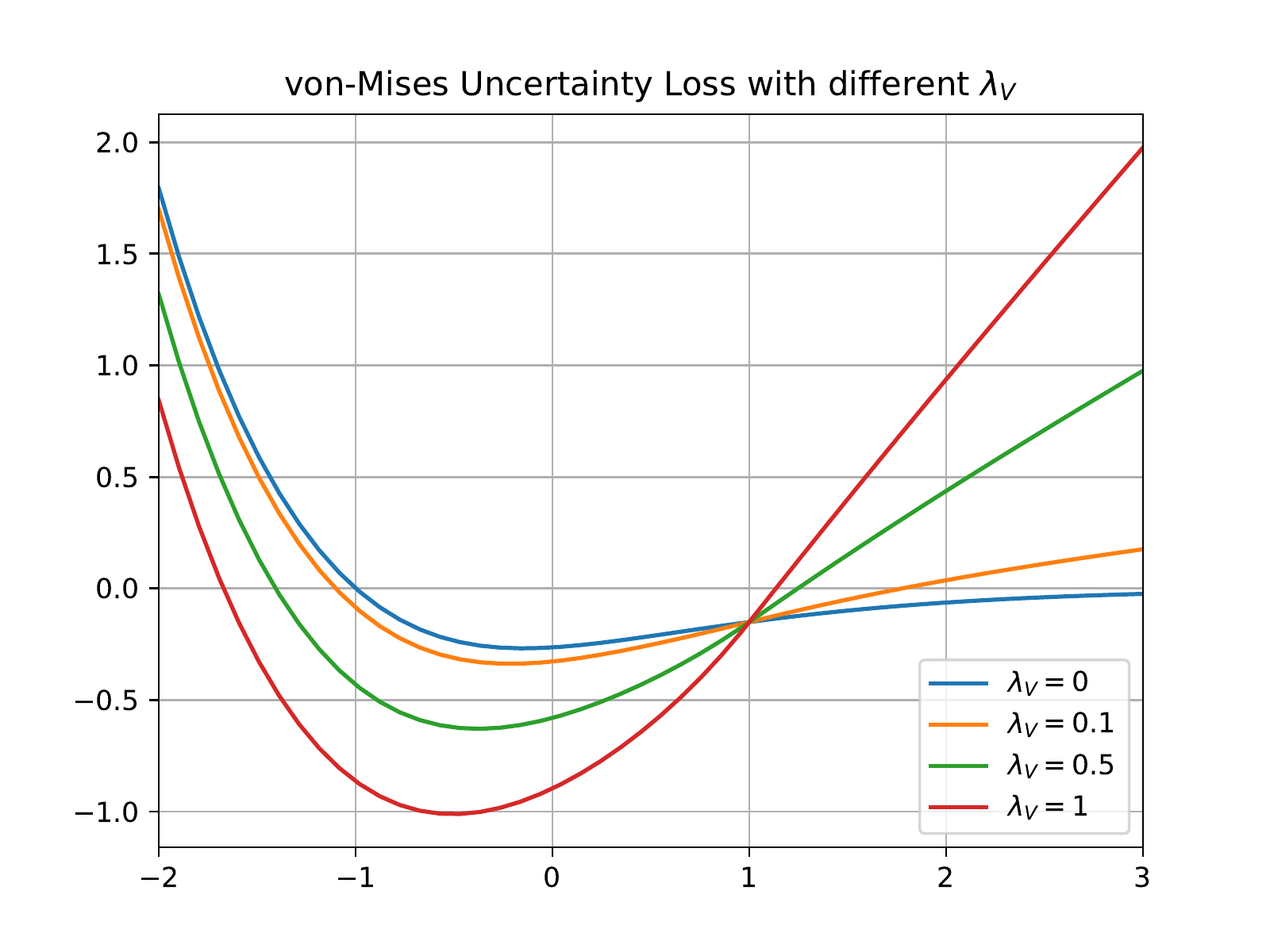}
        \caption{}
    \end{subfigure}
    \caption{Plot (a) illustrates the uncertainty loss functions based on Gaussian distribution with different residuals. (b) shows the Gaussian NLL loss function with different regularization coefficient $\lambda_G$ when $(v-v_t)^2=1$. (c) illustrates the uncertainty loss function based on von-Mises distribution with different residuals, (d) shows the von-Mises NLL loss function with different regularization term $\lambda_V$ when $cos(\theta-\theta_t)=0.5$}
    \label{fig:loss_functions}
    \squeezeuphalf
\end{figure}

\subsubsection{3D Tracking Backbone}

We adopt a detection-based approach similar to \cite{bewley2016simple} with three common components: object association, state estimator and object buffer.

Object association step assigns detections to the existing tracked targets. We use the Hungarian Method to solve the association problem with distance as cost matrix (i.e. use GNN), because in 3D space it's common to find no overlap between two bounding boxes of the same object in adjacent frames. We also set a minimum distance threshold to reject assignments with a larger distance.

The state estimator does filtering on the detection and reports predicted position for association in the next frame. We use an Unscented Kalman Filter with constant turning rate and acceleration assumption on the projected 2D pose ($x,y,\theta$) of the detection. For vertical direction, we just use the lastest $z$ ad $h$ from detection. For box size estimation, we use a simple Kalman Filter to update from detection. Currently for classification, we don't use any filtering and the association will only be performed with objects of the same class.

For the initialization and destruction of the tracked target, we use a simple object buffer strategy. If an object appears in consecutive $t_i$ time, then we consider it to be tracked. If a target has been lost for consecutive $t_d$ time, then we consider the object to be lost.

\subsection{Uncertainty from Heteroscedastic Regression}
To estimate the uncertainty from the network, we added an uncertainty branch after the final RPN feature map. It has a structure similar to the regression branch with the same channel count since we estimate uncertainty for each parameter of a box. Through the heteroscedastic regression, we can directly estimate aleatoric uncertainty while the epistemic uncertainty is ignored. We assume the uncertainties are independent on each box parameter so that we can estimate them separately.

\subsubsection{Regression on linear parameters} We adopt the methodology used in \cite{le2018uncertainty, choi2019gaussian}, i.e. use heteroscedastic regression to estimate the log variance of the parameters. The regression can actually be derived from negative log-likelihood (NLL) of the Gaussian density function. If we assume that each output follows an independent Gaussian distribution, then we can optimize the regression and uncertainty branch by maximizing the probability.
\begin{align*}
    &\arg\max \frac{1}{\sigma\sqrt{2\pi}}\exp\left(-\frac{(v-v_t)^2}{2\sigma^2}\right) \\
    = &\arg\min - \log\left(\frac{1}{\sigma}\exp\left(-\frac{(v-v_t)^2}{2\sigma^2}\right)\right) \\
    = &\arg\min \left(\frac{(v-v_t)^2}{2\sigma^2} + \log\sigma\right) \\
    = &\arg\min \left(\frac{1}{2}\exp(-s)(v-v_t)^2 + \frac{1}{2} s\right)
\end{align*}

Thus the uncertainty loss based on Gaussian distribution can be written as
\begin{gather}
    \label{eq:gaussian_loss} L_G(v,s) = \frac{1}{2} (\exp(-s)(v-v_t)^2 + s)
\end{gather}

Here $v$ is the a parameter of box position or dimension output by the regression branch, and $v_t$ is the encoded ground-truth value. $s:=\log(\sigma^2)$ is the log variance of the parameter $v$, which is the direct output of uncertainty branch. We have a different process method for yaw angle uncertainty, which will be covered by the next subsection.

In \cite{le2018uncertainty}, the authors have discussed why this loss could effective learn the mean and variance at the same time. The loss contains two terms, where the former one is the regression residual with a penalty for lower uncertainty while the latter one could be considered as a regularization term for uncertainty. In Figure \ref{fig:loss_functions}(a) we can also find that the loss functions with different residuals $d^2=(v-v_t)^2$ all have clear minima. A coefficient $\lambda_G$ could be added as follows based on Equation \ref{eq:gaussian_loss} to generalize the regularization. For example, empirical uncertainty bias could be added by using different $\lambda_G$ for objects with different sizes and distances. Figure \ref{fig:loss_functions}(b) prove that we can tune the optimal variance by adjusting the regularization term.
\begin{gather}
    L_G(v,s) = \frac{1}{2} (\exp(-s)(v-v_t)^2 + \lambda_G s)
\end{gather}

\subsubsection{Regression on angular parameters}

The uncertainty loss based on Gaussian distribution should not be applied directly on the yaw angle output, since the angular value is periodic. So we turn to a periodic approximation of Gaussian distribution, which is the von-Mises distribution, to construct a similar uncertainty loss for the rotating angle. The von-Mises distribution has two parameter $\mu,\kappa$ measuring its mean location and concentration respectively. Its probability density function expression is
\begin{gather}
    f_{\text{vM}}(x|\mu,\kappa) = \frac{\exp\left({\kappa\cos(x-\mu)}\right)}{2\pi I_0(\kappa)}
\end{gather}

Here $I_0(\cdot)$ is the 0-order modified Bessel function\footnote{We implemented the modified Bessel autograd function in Pytorch with GPU support. Please refer to \url{https://github.com/cmpute/d3d/tree/master/d3d/math}}. This distribution defines the probability in a periodic way and it satisfies that the integration of the density function on $[-\pi,\pi)$ is 1, which makes it suitable to describe the uncertainty of an angle. Similarly, we create a loss function by maximizing the probability of output angle $\theta$ with target value $\theta_t$ as mean.
\begin{align*}
    &\arg\max f_{\text{vM}}(\theta|\theta_t, \kappa) \\
    = &\arg\max \frac{\exp\left({\kappa\cos(\theta-\theta_t)}\right)}{2\pi I_0(\kappa)} \\
    = &\arg\min -\log\left(\frac{\exp\left({\kappa\cos(\theta-\theta_t)}\right)}{I_0(\kappa)}\right) \\
    = &\arg\min \left(\log I_0(\kappa) - \kappa cos(\theta-\theta_t)\right)
\end{align*}

Since in the von-Mises distribution, the concentration parameter $\kappa$ is actually a reciprocal measure of dispersion, $\kappa$ is analog to $1/\sigma^2$, so in order to have the same definition with linear uncertainty regression and thus apply the uncertainty into tracking module smoothly, we let $s:=-\log\kappa\approx\log(\sigma^2)$, then the loss function for angular value could be written as the following
\begin{gather}
    L_V(v,s) = \log I_0(\exp(-s))-\exp(-s)\cos(\theta-\theta_t)
\end{gather}

This loss function cannot be easily interpreted similarly to the linear version, but according to Figure \ref{fig:loss_functions}(c), we can find that the function still has a clear minimum point when $\cos(\theta-\theta_t) > 0$. However, the loss function doesn't have a minimum when $\cos(\theta-\theta_t)<0$. Aside from that, the gradient of the loss with regard to $s$ is very small when $s$ is large, so we introduced an ELU regularization term to help the optimization without affecting the minimum location a lot. The modified loss function is formulated as follows
\begin{gather}
    \nonumber L_V(v,s) = \log I_0(\exp(-s))-\exp(-s)\cos(\theta-\theta_t) \\
    + \lambda_V ELU(s-s_0)
\end{gather}

Here $\lambda_V$ is the regularization coefficient and $s_0$ controls the position of ELU. Figure \ref{fig:loss_functions}(d) shows the effect of regularization on the loss function when $cos(\theta-\theta_t)=0.5$ and $s_0=1$. From the figure, we can find that the regularization term can help optimization by adding a positive gradient in the right part of the abscissa, and the effect of regularization term on the minimum location could be decreased by adjusting $s_0$.

\subsubsection{Assembling Loss Functions}

With the definition of uncertainty losses, we construct our overall loss function as follows
\begin{gather}
    \nonumber L = \alpha_{cls} L_{cls} + \alpha_{reg} (L_{reg} + \alpha_{angle} L_{reg-\theta}) \\
    + \alpha_{var} (L_{var} + \alpha_{angle} L_{var-\theta})
\end{gather}

Here $L_{cls}$ is the focal loss for classification, $L_{reg}$ is the Smooth-L1 loss for regression, $L_{reg-\theta}$ is the Sine loss for angular regression, $L_{var}$ is the Gaussian uncertainty loss for uncertainty regression, $L_{var-\theta}$ is the von-Mises uncertainty loss for angular uncertainty estimation. $\alpha$ with different subscripts are the weight for the losses respectively. The choose of loss weights will be covered in section \ref{sec:experiments}.

\subsection{Utilization of Uncertainty}

After we gather the uncertainty output from the network, it is important to explain how uncertainty can help object detection and tracking. In this section, we will show how we use uncertainty in the post-processing of the object detection and in the tracking pipeline.

\subsubsection{Scoring with Uncertainty} 


In most object detectors based on RPNs, there is a post-processing setup to remove redundant object proposals. This is usually achieved by Non-Maximum Suppression (NMS), which relies on an object score to filter out objects with lower confidence. It's intuitive to think of adding uncertainty to the score, helping identifying objects with less detection quality. Here we introduce some strategies to incorporate the uncertainty.

Assuming the original detection score (from classification branch or separate objectness branch) is $\beta_d$, then the new score $\beta=\beta_d^\alpha \beta_s^\alpha$. We designed three functions to map log uncertainty $s$ to log score $\log(\beta_s)$ with a scaling parameter $k_s>0$ and an offset parameter $b_s>0$. The general idea here is to let the box with lower uncertainty have a higher score.

\begin{itemize}
    \item Linear: $\log(\beta_s) = \max\{-k_s g(s)+b_s, 0\} (k_s>0)$
    \item Exponential: $\log(\beta_s) = -\exp(k_s g(s)+b_s)$
    \item Sigmoid: $\log(\beta_s) = \log(\text{sigmoid}(-k_s g(s)+b_s))$
\end{itemize}

Here $g(s)$ is an aggregation function of $s$ from different box parameters, common options including maximum and summation. These mapping functions are designed empirically and we would discuss the effect of these functions in the Section \ref{sec:experiments}. We use mapping from logarithm to logarithm to avoid possible numerical issues.

\begin{table*}[t]
    \centering
    \captionsetup{justification=centering}
    \caption{3D Object (Cars) Detection Performance Comparison on KITTI object (val) dataset. \\ Regression column: B - SmoothL1 box regression; G - Gaussian variance regression; V - von-Mises variance regression. \\ The \textit{Best Settings} row shows the best performance among all our experiments in Table \ref{tab:params_ablation}}
    \footnotesize
    \begin{tabular}{c|l|c|l|ccc|ccc}
        \toprule
        \multirow{2}{*}{} & \multirow{2}{*}{Method} & \multirow{2}{*}{Source} & \multirow{2}{*}{Regression} & \multicolumn{3}{c|}{$\text{AP}_{3D}\% (IoU>0.7) \uparrow$}  & \multicolumn{3}{c}{$\text{AP}_{BEV}\% (IoU>0.7) \uparrow$} \\
        & & & & Easy & Moderate & Hard & Easy & Moderate & Hard \\
        \midrule
        \multirow{7}{*}{\begin{sideways}Literature\end{sideways}}
        & MV3D\cite{chen2017multi}          & \cite{sindagi2019mvx}  & B & 71.3 & 62.7 & 56.6 & 86.6 & 78.1 & 76.7 \\
        & PIXOR\cite{yang2018pixor}          & \cite{sindagi2019mvx}  & B & -    & -    & -    & 86.8 & 80.8 & 76.6 \\
        & F-PointNet\cite{qi2018frustum}    & \cite{sindagi2019mvx}  & B & 83.8 & 70.9 & 63.7 & 88.2 & 84.0 & 78.6 \\
        & VoxelNet\cite{zhou2018voxelnet}   & \cite{sindagi2019mvx}  & B & 82.0 & 65.5 & 62.9 & 89.6 & 84.8 & 78.6 \\
        & SECOND\cite{yan2018second}        & \cite{he2020structure} & B & 87.43 & 76.48 & 69.10 & - & - & - \\
        & PointRCNN\cite{shi2019pointrcnn}  & \cite{he2020structure} & B & 88.88 & 78.63 & 77.38 & - & - & - \\
        & SA-SSD\cite{he2020structure}      & \cite{he2020structure} & B & 90.15 & 79.91 & 78.78 & - & - & - \\
        
        \midrule
        \midrule

        \multirow{5}{*}{\begin{sideways}Experiments\end{sideways}}
        & SECOND (Reproduced Baseline)              & - & B & 87.83 & 78.05 & 76.62 & 89.66 & 87.77 & \textbf{86.04} \\
        \cmidrule{2-10}
        & a) SECOND (Default Settings)                 & - & B+GV & 88.87 & \textbf{78.77} & \textbf{77.86} & 90.09 & \textbf{87.76} & 84.79 \\ 
        & b) SECOND (Variance Only)                    & - & GV & 86.78 & 76.62 & 74.21 & 89.21 & 85.12 & 79.11 \\ 
        & c) SECOND (Gaussian Only)                    & - & B+G & \textbf{88.93} & 78.55 & 77.64 & \textbf{90.23} & 87.36 & 84.15 \\ 
        & -) \textit{SECOND (Best Settings)}          & - & B+GV & \textit{89.30} & \textit{79.34} & \textit{78.61} & \textit{90.35} & \textit{88.50} & \textit{87.50} \\ %

        \bottomrule
    \end{tabular}
    \label{tab:detection_results}
\end{table*}

\subsubsection{Tracking with Uncertainty}

Before using the uncertainty, we need to note that the output of the network is the log uncertainty $s=\log(\sigma^2)$ of the encoded box parameters. So in order to get the actual uncertainty we need to refer to the box parameter encoding and transform $s$ accordingly.
\begin{gather}
    x_t=\frac{x-x_a}{d_a} \Rightarrow \mathbb{V}[x]=d_a^2\mathbb{V}[x_t] \\
    y_t=\frac{y-y_a}{d_a} \Rightarrow \mathbb{V}[y]=d_a^2\mathbb{V}[y_t] \\
    z_t=\frac{z-z_a}{h_a} \Rightarrow \mathbb{V}[z]=h_a^2\mathbb{V}[z_t] \\
    \label{eq:uncertainty_w} w_t=\log\left(\frac{w}{w_a}\right) \Rightarrow \mathbb{V}[w]\approx\mathbb{E}^2[w]\mathbb{V}[w_t] \\
    \label{eq:uncertainty_l} l_t=\log\left(\frac{l}{l_a}\right) \Rightarrow \mathbb{V}[l]\approx\mathbb{E}^2[l]\mathbb{V}[l_t] \\
    \label{eq:uncertainty_h} h_t=\log\left(\frac{h}{h_a}\right) \Rightarrow \mathbb{V}[h]\approx\mathbb{E}^2[h]\mathbb{V}[h_t] \\
    \theta_t = \theta-\theta_g \Rightarrow \mathbb{V}[\theta]=\mathbb{V}[\theta_t]
\end{gather}

\begin{table}[t]
    \centering
    \captionsetup{justification=centering}
    \caption{3D Object (Cars) Detection Performance Comparison with different variance regression parameters.}
    \resizebox{\columnwidth}{!}{
    \footnotesize
    \begin{tabular}{c|cccc|ccc|ccc}
        \toprule
        \multirow{2}{*}{\#} & \multirow{2}{*}{$\alpha_{var}$} & \multirow{2}{*}{$\alpha_{angle}$} & \multirow{2}{*}{$\lambda_G$} & \multirow{2}{*}{$\lambda_V$} & \multicolumn{3}{c|}{$\text{AP}_{3D}\% (IoU>0.7) \uparrow$}  & \multicolumn{3}{c}{$\text{AP}_{BEV}\% (IoU>0.7) \uparrow$} \\
        & & & & & Easy & Moderate & Hard & Easy & Moderate & Hard \\
        \midrule
        a) & 1 & 1 & 1 & 1 & 88.87 & 78.77 & 77.86 & 90.09 & 87.76 & 84.79 \\ 
        \midrule
        d) & 1 & 4 & 1 & 1 & 88.68 & 78.89 & 78.31 & 89.99 & 88.08 & \textbf{87.50} \\ 
        e) & 1 & 1 & 0.5 & 1 & \textbf{89.30} & \textbf{79.34} & \textbf{78.61} & 90.22 & 88.32 & 87.45 \\ 
        f) & 1 & 1 & 1 & 0.5 & 88.52 & 78.40 & 77.39 & 89.99 & 87.96 & 84.96 \\ 
        g) & 0.5 & 1 & 1 & 1 & \textbf{89.29} & 79.26 & 78.37 & \textbf{90.35} & 88.31 & 86.90 \\ 
        h) & 0.2 & 1 & 1 & 1 & 89.22 & 79.26 & 78.20 & 90.21 & \textbf{88.50} & 87.39 \\ 
        \bottomrule
    \end{tabular}
    }
    \label{tab:params_ablation}
\end{table}

\begin{table}[t]
    \centering
    \captionsetup{justification=centering}
    \caption{3D Object Detection Performance Comparison with different NMS strategies on KITTI object (val) dataset.\\ C stands for NMS using classification score, L/E/S stands for linear, exponential, sigmoid mapping of uncertainty.}
    \resizebox{\columnwidth}{!}{
    \footnotesize
    \begin{tabular}{l|ccc|ccc}
        \toprule
        \multirow{2}{*}{NMS Strategy} & \multicolumn{3}{c|}{$\text{AP}_{3D}\% (IoU>0.7) \uparrow$}  & \multicolumn{3}{c}{$\text{AP}_{BEV}\% (IoU>0.7) \uparrow$} \\
        & Easy & Moderate & Hard & Easy & Moderate & Hard \\
        \midrule
        a) C (baseline)               & 88.87 & 78.77 & 77.86 & 90.09 & 87.76 & 84.79 \\
        \midrule
        i) C+L ($k_s=0.001, b_s=0$)  & \textbf{88.89} & 78.77 & 77.86 & 90.09 & 87.77 & 84.80 \\
        j) C+L ($k_s=0.01, b_s=0$)   & 88.79 & 78.73 & 77.85 & 90.04 & 87.77 & 84.80 \\
        k) C+S ($k_s=0.001, b_s=0$)  & 88.88 & 78.77 & 77.86 & 90.09 & 87.76 & 85.39 \\
        l) C+S ($k_s=0.01, b_s=0$)   & 88.86 & 78.77 & 77.86 & 90.07 & \textbf{87.78} & 85.40 \\
        m) C+E ($k_s=0.001, b_s=0$)  & \textbf{88.89} & 78.77 & 77.86 & 90.09 & 87.77 & 86.12 \\
        n) C+E ($k_s=0.01, b_s=0$)   & 88.79 & 78.73 & 77.85 & 90.04 & 87.77 & \textbf{86.16} \\
        \bottomrule
    \end{tabular}
    }
    \label{tab:nms_comparison}
\end{table}

\begin{table*}[t]
    \centering
    \captionsetup{justification=centering}
    \caption{3D Object (Cars) Tracking Performance Comparison on KITTI tracking dataset. \\ The threshold for a true positive match is IoU > 0.5 \\ In the baseline method, $\sigma$=1 means using an identity matrix as the observation noise covariance matrix, $\sigma$=median means using the median of our uncertainty estimation as the covariance matrix diagonal values.}
    \footnotesize
    \begin{tabular}{c|l|l|cc|cccc}
        \toprule
        \# & Method & Regression & AP\%$\uparrow$ & Max F1\%$\uparrow$ & IDSW$\downarrow$ & FRAG$\downarrow$ & ML\%$\downarrow$ \\
        \midrule
        & Constant $\sigma$=1              & B    & 13.0 & 30.9 & 39517 & 30934 & 38.9 \\
        & Constant $\sigma$=median         & B    & 29.4 & 48.4 & 41138 & 30254 & 27.8 \\
        \midrule
        a) & Default Settings                 & B+GV & 29.1 & 45.0 & 23309 & \textbf{18597} & 52.8 \\ 
        b) & Variance Only                    & GV   & \textbf{42.3} & \textbf{55.1} & 35937 & 28693 & 30.6 \\ 
        c) & Gaussian Only                    & B+G  & 20.7 & 39.6 & \textbf{21566} & 18846 & 41.7 \\ 
        d) & $\alpha_{angle}=4$               & B+GV & 30.5 & 46.5 & 30265 & 21161 & 38.9 \\ 
        e) & $\lambda_G=0.5$                  & B+GV & 27.4 & 44.5 & 26956 & 24983 & 41.7 \\ 
        f) & $\lambda_V=0.5$                  & B+GV & 26.3 & 40.6 & 33997 & 25557 & 36.1 \\ 
        g) & $\alpha_{var}=0.5$               & B+GV & 34.1 & 47.1 & 35091 & 26843 & \textbf{25.0} \\ 
        h) & $\alpha_{var}=0.2$               & B+GV & 27.4 & 46.6 & 35374 & 32203 & 27.8 \\ 
       
        \bottomrule
    \end{tabular}
    \label{tab:tracking_results}
    \squeezeup
\end{table*}

The approximation of Equation \ref{eq:uncertainty_w} to \ref{eq:uncertainty_h} is based on the Taylor expansion of the expectation and variance of a random variable \cite{benaroya2005probability}. Given a random variable $\mathcal{X}$ and function $f(\mathcal{X})$, with Taylor expansion of $f(\mathcal{X})=f(\mathbb{E}[\mathcal{X}])+(\mathcal{X}-\mathbb{E}[\mathcal{X}])$ we can get $\mathbb{V}[f(\mathcal{X})]\approx \left(f'(\mathbb{E}[\mathcal{X}])\right)^2\sigma^2_\mathcal{X}$. So take $w$ for example, $\mathbb{V}[w_t]=\mathbb{V}[\log(\frac{w}{w_a})]=\left(\mathbb{E}^{-1}[\frac{w}{w_a}]\right)^2\mathbb{V}[\frac{w}{w_a}]=\mathbb{E}^{-2}[w]\mathbb{V}[w]$.

After the estimation of the uncertainties, we can use the diagonal uncertainty matrix (filled by the uncertainty values from the network) as the covariance of initial state noise and the covariance of observation noise afterward. These is very different from \cite{bewley2016simple} which assumes constant covariance values for initial state noise, process noise and observation noise. Nevertheless, we still leave the covariance matrix of process noise as a hyperparameter to be tuned.

\section{Experiment Results}
\label{sec:experiments}

In this section, we are going to demonstrate the effectiveness of additional uncertainty estimation by experiments of object detection and tracking with datasets.

\subsection{Training Details}

The backbone detection network we use is the SECOND \cite{yan2018second} as mentioned before and we added the uncertainty branch after the Region Proposal Network. We also implemented the tracking pipeline in our code base. Instead of using any pre-trained weights, we train the networks from scratch for 40 epochs on the training split\cite{chen2017multi} of the KITTI object dataset\cite{geiger2013vision}. After the split, the KITTI training set contains 3712 frames and the validation set consists of 3769 frames. The network is built with Pytorch and trained on 2 NVIDIA RTX2080s using the Adam optimizer. After training, the object detection performance is evaluated on the validation set of KITTI object dataset, while the object tracking performance is evaluated on KITTI tracking dataset.

In default settings, we use Focal Loss with $\gamma=2$ as the classification loss and SmoothL1 Loss as the box regression loss as mentioned in Section \ref{sec:methodology}. We set the weights as $\alpha_{cls}=1$, $\alpha_{reg}=2$, $\alpha_{angle}=1$. The default settings in uncertainty branch are $\lambda_G=\lambda_V=1$. For detection post-processing, we use normal NMS to filter out redundant boxes after selecting the 100 boxes with the highest scores.

\subsection{Object Detection Performance}

We use average precision, a common object detection evaluation metric, to compare the performance of our model versus baseline models and we display AP of different types of objects as designed by KITTI. 

Table \ref{tab:detection_results} listed the detection performance of our method compared with other methods. As shown in the figure, our network with the uncertainty branch outperforms the original method on the detection performance in most metrics and there's not prominent degradation in other metrics. This proves that our method handles the detection better with uncertainty estimated.

\subsection{Object Tracking Performance}

For the evaluation of object tracking performance, we also inherit the common metrics from popular benchmarks like KITTI. These metric includes average precision (AP), maximum F1 score (Max F1), id switches (IDSW), fragments (FRAG) and mostly lost ratio (ML). The baseline method is using the original SECOND network without uncertainty output, and constant variance values of observation noise are used for each box parameter.

The results in Table \ref{tab:tracking_results} show that our modification on the classical Kalman Filter based tracking framework can improve the tracking performance by a large margin, thanks to more reasonable uncertainty input to the tracker. Comparing the tracking performance of two cases with constant covariance matrices, we can also find that the Kalman Filter is sensitive to the covariance matrix settings. So our method still gives a reasonable reference value if a constant covariance matrix has to be applied.

\subsection{Ablation Study}

\subsubsection{Uncertainty Loss Weight and Regularization}

Specifying the weight of the uncertainty loss and coefficient of regularization both makes a difference to the result. From Table \ref{tab:detection_results} and \ref{tab:params_ablation}, we can compare (a)(b)(c) to find that although similar performance can be achieved using only uncertainty regression, combine original box regression and both variance regression methods we can get better detection performance. Compare (a)(g)(h) we can find that large $\alpha_{var}$ will lead to performance decline. Compare (a)(e)(f) we can find that regularization coefficients can make a difference to the performance, yet a general paradigm for tuning the coefficients is not clear.

\subsubsection{Scoring with Uncertainty mapping}

Taking uncertainty into account during the NMS process could help to choose boxes with higher localization quality in the NMS process, as proved by experiments shown in Table \ref{tab:nms_comparison}. The largest improvement comes from the exponential mapping in terms of $\text{AP}_{BEV}$ of hard samples. Although the overall improvement is not huge, and the selection of mapping strategy and the coefficients are empirical, they can be considered as hyper-parameters that are easy to tune. A simple grid search of the uncertainty mapping and its coefficient can be performed after the training of the network. 


\section{Conclusion}
\label{sec:conclusion}

In this paper, we introduced the aleatoric uncertainty estimation branch to the 3D object detector and created a novel von-Mises NLL loss for uncertainty estimation of periodic parameters. The detector performs better with the uncertainty estimation. We also feed the uncertainty into the 3D object tracking framework which is an important use case for the estimated uncertainty. The effectiveness of the adaptive uncertainty input in tracking is proved by experiments. Another use case for the estimated uncertainty is generating additional scores for the NMS process in object detection. By this combination, we created a novel and low-cost way to finetune the performance of object detectors.

Although most of our experiments are conducted based on the SECOND framework, our method can be applied to any 3D object detectors with a box parameter regression branch by adding a similar branch with the uncertainty losses.

The proposed algorithms are early attempts to add uncertainty to deep learning based 3D object detection and tracking framework. Further exploration directions include using the Kalman filter in object classification with proper distribution assumption (such as Dirichlet distribution) and removing the independence assumption of the box parameters by direct estimate the whole covariance matrix.

\addtolength{\textheight}{-7.5cm}   





\bibliographystyle{IEEEtran}
\bibliography{ref}

\end{document}